\documentclass[runningheads]{llncs}
\usepackage{graphicx}
\usepackage{epstopdf}
\usepackage{caption}
\usepackage{subfigure}
\usepackage{color}
\usepackage{ulem}
\usepackage{textcomp}
\usepackage{tipa}
\usepackage{amssymb}
\usepackage{dsfont}
\usepackage{subfigure}
\usepackage{mathrsfs}
\usepackage{bm}
\usepackage{multirow}
\usepackage{amsmath}
\usepackage{cite}
\usepackage[colorlinks,linkcolor=blue,anchorcolor= blue,citecolor=blue]{hyperref}
\usepackage{comment}
\usepackage{booktabs}
\usepackage{array}
\usepackage{comment}
\usepackage{xcolor}
\newcolumntype{P}[1]{>{\centering\arraybackslash}p{#1}}
\newcolumntype{M}[1]{>{\centering\arraybackslash}m{#1}}
\newcommand{\mb}{\mathbf}
\newcommand{\samethanks}[1][\value{footnote}]{\footnotemark[#1]}
\begin{document}
\title{Superpixel-guided Iterative Learning from Noisy Labels for Medical Image Segmentation}
\titlerunning{Superpixel-guided Iterative Learning from Noisy Labels}

\author{Shuailin Li \thanks{Equal contribution. This work was supported by Shanghai Science and Technology Program 21010502700 and by the ShanghaiTech-UII Joint Lab.} \inst{1} \and 
	Zhitong Gao \samethanks \inst{1} \and
	Xuming He\inst{1,2}}

\institute{
	ShanghaiTech University, Shanghai, China \\ \and
	Shanghai Engineering Research Center of Intelligent Vision and Imaging \\
	\email{\{lishl, gaozht, hexm\}@shanghaitech.edu.cn}
}

%
\maketitle              

\begin{abstract}
{
Learning segmentation from noisy labels is an important task for medical image analysis due to the difficulty in acquiring high-quality annotations.
Most existing methods neglect the pixel correlation and structural prior in segmentation, often producing noisy predictions around object boundaries. 
To address this, we adopt a superpixel representation and develop a robust iterative learning strategy that combines noise-aware training of segmentation network and noisy label refinement, both guided by the superpixels. This design enables us to exploit the structural constraints in segmentation labels and effectively mitigate the impact of label noise in learning.
Experiments on two benchmarks show that our method outperforms recent state-of-the-art approaches, and achieves superior robustness in a wide range of label noises.
Code is available at 
\href{https://github.com/gaozhitong/SP_guided_Noisy_Label_Seg}{https://github.com/gaozhitong/SP\_guided\_Noisy\_Label\_Seg}.
}
{
}

\end{abstract}

\section{Introduction}

Semantic segmentation of medical images, a fundamental task in computer-aided clinical diagnoses, has recently achieved remarkable progress thanks to the effective feature learning based on deep neural networks (DNN)~\cite{Ching2018OpportunitiesAO,Litjens_2017,Shen2017DeepLI}. Training such DNNs for segmentation typically requires a large dataset with pixelwise annotations that accurately delineate object boundaries. In the medical domain, however, acquiring such high-quality annotations is often difficult due to lack of experienced annotators and/or visual ambiguity in object boundaries~\cite{Kohli2017MedicalID, 7046014}. Consequently, the annotated datasets often include a varying amount of label noise in practice, ranging from small boundary offsets to large region errors. Learning from those noisy annotations has been particularly challenging for deep segmentation networks due to the memorization effect~\cite{arpit2017closer}.   
 
There have been several attempts to tackle the problem of training segmentation networks from noisy labels, which can be largely grouped into two categories. The first type of methods view the annotation of each image as either clean or corrupted, and iteratively select or reweight image samples during training~\cite{Zhu2019PickandLearnAQ, Xue2020CascadedRL}. In particular, Zhu et al.~\cite{Zhu2019PickandLearnAQ} implicitly reweight image losses by simultaneously training a label evaluation network and a segmentation network, while Xue et al.~\cite{Xue2020CascadedRL} explicitly select a subset of images by extending the Co-teaching~\cite{Han2018CoteachingRT} scheme into a tri-network framework. Such image-level weighting strategies, however, are less robust under severe noise settings as they are unable to fully utilize the pixels with clean annotations in each image. To address this limitation, the second group of training methods consider the segmentation as a pixel-wise classification task~\cite{Zhang2020CharacterizingLE, Zhang2020RobustMI}, and perform pixel-wise sample selection or label refinement based on the state-of-the-art robust classifier learning strategies, such as the confidence learning technique~\cite{Northcutt2019ConfidentLE} and the Co-teaching method with tri-networks~\cite{Zhang2020RobustMI}. Despite their better use of annotations, the pixel-level approaches ignore the pixel correlation and spatial prior in image segmentation, and hence tend to produce noisy prediction around object boundaries. 

In this work, we propose a novel robust learning strategy for semantic image segmentation, aiming to exploit the structural prior of images and correlation in pixel labels. To this end, we adopt a superpixel representation and develop an iterative learning scheme that combines noise-aware training of segmentation network and noisy label refinement, both guided by the superpixels. Such integration allows us to better utilize the structural constraint in segmentation labels for model learning, which can effectively mitigate the impact of label noise.  We note that while superpixel has been employed in recent work~\cite{li2019supervised}, they only use it to correct noisy labels and ignore the impact of noise during training.

Specifically, in each iteration, we first jointly train two deep networks using selected subsets of superpixels with small loss values, following the multi-view learning framework~\cite{Han2018CoteachingRT,Wei2020CombatingNL}. As in the Co-teaching method, such a multi-view learning strategy regularizes the network training via the predictions of the peer networks. Here we treat each superpixel as a data sample in selection, which enables us to enforce spatial smoothness and provide better object boundary cues in network training. To avoid overfitting to label noise, we design an automatic stopping criterion for the joint learning based on the loss statistics of superpixels. After the network training, we then use the network predictions to estimate the reliability of superpixel labels and relabel a subset of most unreliable ones. Such label refinement allows us to improve the label quality for the subsequent model training. The network and label updates are repeated until no further improvement can be achieved for the label refinement.      

We evaluate our method on two public benchmarks, ISIC skin lesion dataset~\cite{Gutman2018SkinLA} and JSRT chest x-ray dataset~\cite{Shiraishi2000DevelopmentOA,Ginneken2006SegmentationOA}, under extensive noise settings. Empirical results show that our method consistently outperforms the previous state of the art and demonstrates training robustness in a wide range of label noises. 

\section{Method}
\begin{figure*}[t!]
	\centering 
	\includegraphics[width=1.0\textwidth]{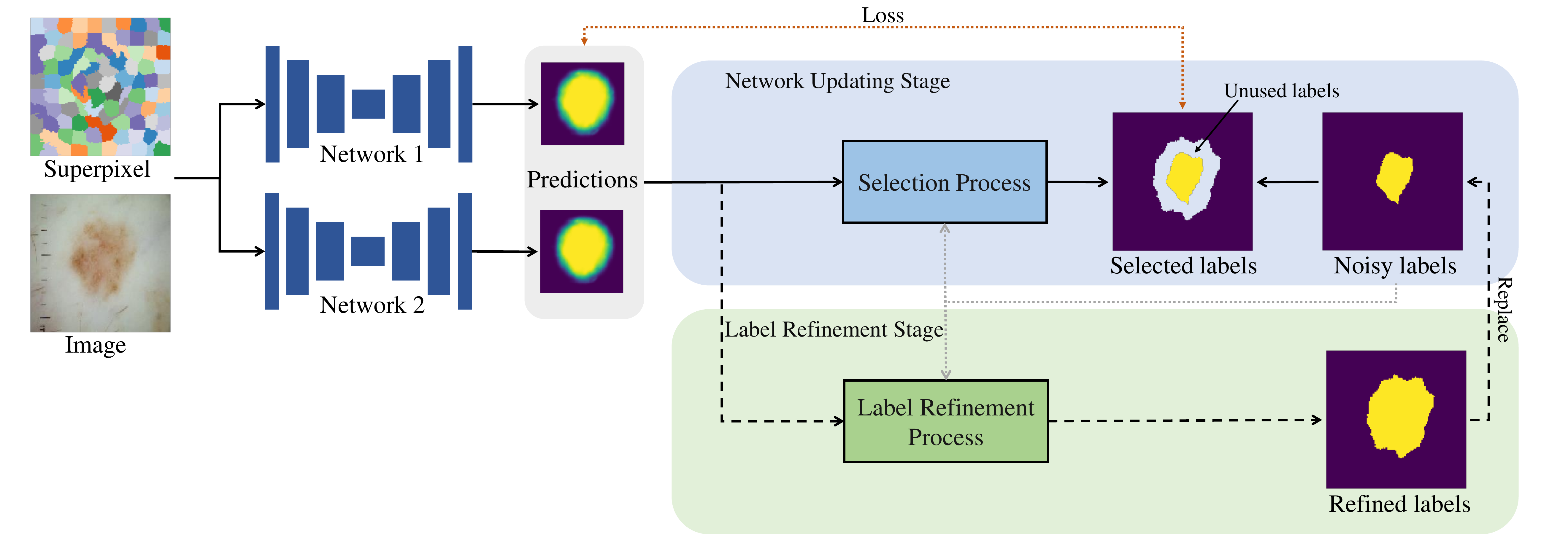}
	\caption{Overview of our robust training process. We use superpixels as our guidance in an iterative learning process which jointly updates network parameters and refines noisy labels. Each iteration selects superpixels with small losses to update two networks and relabels a set of superpixels based on network outputs.} 
	\label{Fig:overview}
\end{figure*}


We now introduce our robust learning strategy for semantic segmentation, which aims to exploit the structural constraints in the label masks and to fully utilize reliable pixel-level labels for effective learning. To achieve this, we adopt a superpixel-based data representation, and develop an iterative learning method that jointly optimizes the network parameters and refines noisy labels.  

Specifically, given a target network and noisy training data, we first compute the superpixels of the input images. 
Based on such pixel groupings, our iterative learning procedure alternates between a noise-aware network training stage and a label refinement stage until no improvement can be achieved. For the network training stage, we adopt the multi-view learning framework which jointly trains two instances of the segmentation network. 
For the label refinement stage, we use the output of two trained networks to estimate the reliability of superpixel labels and to update the unreliable labels. An overview of our training pipeline is shown in Figure~\ref{Fig:overview}. Below we first present our superpixelization procedure in Sec.~\ref{sec_perceptual}, followed by the two stages in the iterative learning in Sec.~\ref{Sec:Network Updating}.

\subsection{Superpixel Representation}\label{sec_perceptual}  
To exploit the image structural prior and spatial correlation in pixel labels, we first compute a superpixel representation for the training images. Such superpixel representation has been shown effective for different medical image modalities in literature, e.g. ~\cite{qin2018superpixel} for CT, ~\cite{tian2015superpixel} for MR and ~\cite{daoud2019automatic} for US images. Specifically, we use the off-the-shelf superpixelization method, SLIC~\cite{Achanta2012SLICSC}, to partition each image into a set of homogeneous regions. For color images, we adopt the CIE-lab color space to represent pixel features, while for other modalities, such as X-ray images, we use both the pixel intensity and deep features from a U-net trained with a noise-aware method, e.g.~\cite{Wei2020CombatingNL}.
\footnote{On the chest x-ray dataset~\cite{Shiraishi2000DevelopmentOA,Ginneken2006SegmentationOA}, our superpixels achieve undersegmentation errors lower than 0.32 for 800 superpixels per image, which is comparable to the natural image setting\cite{Achanta2012SLICSC}.}

We assume the pixels share similar groundtruth labels in each superpixel, which enable us to enforce the structural constraints on the label masks and better preserve object boundaries. More importantly, we treat each superpixel as a data sample in the subsequent robust network learning as well as the label refinement. This allows us to estimate the noise level of pixel annotations in a more reliable manner by pooling the pixel information from each superpixel.

\subsection{Iterative Model Learning}\label{sec_iter}
We now present our iterative learning strategy based on the superpixel representation, which aims to fully utilize the clean pixel annotations and meanwhile reduce the impact of noisy labels. To this end, we introduce an iterative optimization process for model training as below. Each iteration consists of two stages: a noise-aware network learning stage to update the network parameters and a label refinement stage to correct unreliable annotations.   

\subsubsection{Network Update}\label{Sec:Network Updating}

In the first stage, we perform a noise-aware network learning by incorporating superpixel representation into a multi-view learning framework. Specifically, following the Co-teaching strategy~\cite{ren2018learning, jiang2018mentornet, Han2018CoteachingRT}, we jointly train two instances of the target segmentation network using partial data with small losses~\cite{arpit2017closer}. To better select data samples with clean labels, we design a superpixel-wise loss that combines the loss values of two networks with an agreement-based regularization~\cite{Wei2020CombatingNL} on superixels. Our loss provides a reliable guidance to sample selection thanks to the structural prior encoded in the superpixels.  

Formally, given an image $\mb{X}$, we denote its annotation as $\mb{Y}=\{Y_i\}_{i=1}^M, Y_i\in\{1,\cdots, C\}$ where $C$ is the number of semantic classes and $M$ is the number of pixels. The superpixel map is represented by $\mb{S} =\{S_i\}_{i=1}^M $ where $S_i\in\{1,2,\cdots, K\}$ and $K$ is the number of superpixels. Here $S_j=k$ means that pixel $j$ belongs to superpixel $k$.

We aim to train two deep neural networks denoted by $f(\cdot, \theta_1)$ and $f(\cdot, \theta_2)$. 
To define a loss for each image, we first generate the predicted probability maps from two networks, denoted by $\mb{P}^1,\mb{P}^2\in \mathbb{R}^{C \times M}$, where $\mb{P}^i=f(\mb{X}, \theta_i), i=1,2$. We then compute the superpixel-wise probabilities  $\mb{P}_s^i\in\mathbb{R}^{C \times K}, i=1,2$ and the corresponding soft labels ${\mb{Y}_s}\in[0,1]^{C\times K}$ by averaging over each superpixel:
\begin{equation}\label{eq:joint_prob}
\mb{P}^i_s(c,k) = \frac{1}{N(k)}\sum_{j:S_j = k} \mb{P}^i(c,j), \qquad 
\mb{Y}_s(c,k) =\frac{1}{N(k)} \sum_{j:S_j = k}\mathds{1}(Y_j = c)
\end{equation}
where $N(k) = |\{j:S_j = k\}|$ is the size of the superpixel.
Inspired by \cite{Wei2020CombatingNL}, we define our superpixel-wise loss function $\ell^{sp}$ by considering both classification losses and prediction agreement on each superpixel:
\begin{equation}\label{l_k}
	\ell^{sp} = (1-\lambda)*(\ell_{ce}(\mb{P}^1_s,\mb{Y}_s) + \ell_{ce}(\mb{P}^2_s,\mb{Y}_s)) + \lambda *\ell_{kl}(\mb{P}^1_s,\mb{P}^2_s)
\end{equation}
where $\ell_{ce}$ is the cross-entropy loss with soft labels, $\ell_{kl}$ is the symmetric Kullback-Leibler(KL) Divergence, and $\lambda $ is a balance factor. By considering both two terms in the small loss criterion, we aim to select and update on training data with low label noise while maximizing networks' agreement. 
Denote $R$ as the ratio of pixels being selected, we perform the small-loss selection by choosing the superpixel set $\mathcal{\hat{D}}_s$ as follows:
\begin{equation}
	{\mathcal{\hat{D}}}_s = {\arg\min}_{{\mathcal{D}_s}: N({\mathcal{D}_s})\ge R\cdot M} \sum_{k\in{\mathcal{D}_s}}\ell^{sp}_k 
\end{equation}
where $N({\mathcal{D}_s}) = \sum_{k\in {\mathcal{D}_s}} N(k)$ is the total number of pixels in the superpixel set. Given the small-loss selection, we train the network based on the average loss:
\begin{equation}
	\mathcal{L} = \frac{1}{N(\mathcal{\hat{D}}_s)}\sum_{S_i \in \mathcal{\hat{D}}_s}\ell_i
\end{equation}
where $\ell$ has the same form as Eq.~\ref{l_k} except it is defined on the pixel level. Here we skip the superpixel-level pooling as in Eq.~\ref{eq:joint_prob} for more efficient back propagation. %

\subsubsection{Stopping Criterion}
While the selection strategy enables the network training with mostly clean-labeled data, some noisy labels are inevitably selected and gradually affect model performance. To tackle the problem, we propose a criterion to stop network training before such overfitting. Our criterion is defined based on the loss gap $G_l$ between the selected data and the rest of the training set as follows:
\begin{equation}\label{loss_gap}
	G_{l} =    \frac{1}{K-|\mathcal{\hat{D}}_s|}\sum_{k \notin \mathcal{\hat{D}}_s}\ell_k^{sp} - \frac{1}{|\mathcal{\hat{D}}_s|}\sum_{k \in \mathcal{\hat{D}}_s}\ell_k^{sp} 
\end{equation}
Intuitively, the model tends to first learn the relatively simple patterns in clean data, then starts to overfit to label noise~\cite{arpit2017closer}. Consequently, $G_{l}$ first gradually increases, then starts to decrease. Based on this observation, we stop the model training when $G_{l}$ reached the maximum before decreasing.

After training, we observe that the outputs of two peer networks are very similar to each other. Consequently, we arbitrarily choose one network to make predictions during test/deployment.


\subsubsection{Label Refinement}
In this stage, we use the trained networks to estimate the reliability of the superpixel annotations and relabel a subset of unreliable ones. Specifically, we choose superpixels with large losses, which indicates strong inconsistency between model predictions and their labels. We then relabel them according to predicted class labels.  
Formally, we define the unreliable superpixel set $\mathcal{\hat{D}}_u$ based on the superpixel losses and compute the predicted superpixel labels $\mb{\hat{Y}} =\{\hat{Y}_i\}_{i=1}^K,\hat{Y}_i\in \{1,\cdots,C\}$ as below, 
\begin{align}
{\mathcal{\hat{D}}}_u &= {\arg\max}_{{\mathcal{D}_u}: N({\mathcal{D}_u})\le (1-{R})\cdot M} \sum_{k\in{\mathcal{D}_u}}\ell_k^{sp} \\
\hat{Y}_k &= \mathop{\arg\max}_{c}\frac{1}{2}(\mathbf{P}_s^1(c,k) + \mathbf{P}_s^2(c,k))
\end{align}
where $\ell_k^{sp}$ is defined in Eq.~\ref{l_k}, $R$ is the selection ratio mentioned before.  Finally, we update the pixel-wise label map $\mb{Y}'=\{Y'_i\}_{i=1}^M, Y'_i\in\{1,\cdots, C\}$ as
\begin{equation}
	Y_i' = \mathds{1}(S_i \in {\mathcal{\hat{D}}}_u )\hat{Y}_{S_i} + \mathds{1}(S_i \notin {\mathcal{\hat{D}}}_u )Y_i
\end{equation}
After the label refinement, we replace $\mb{Y}$ by $\mb{Y}'$, increase $R$ by a fixed ratio $\gamma$ and start the next iteration.

\section{Experiment} \label{sec:exp}
We validate our method on two public datasets, ISIC~\cite{Gutman2018SkinLA} and JSRT~\cite{Shiraishi2000DevelopmentOA,Ginneken2006SegmentationOA}, which consists of images from two different modalities. We follow the literature and use simulated label noises as no public benchmark with real label noises is available.


\subsection{Dataset}
\textbf{ISIC Dataset.} ISIC 2017 dataset~\cite{Gutman2018SkinLA} is a public large-scale dataset of dermoscopy images, acquired from a variety of devices used at multiple sites. This dataset contains 2000 training and 600 test images with corresponding segmentation masks. We resize all images to 128$\times$128 in resolution.

\noindent\textbf{JSRT Dataset.} JSRT dataset~\cite{Shiraishi2000DevelopmentOA,Ginneken2006SegmentationOA} is a public chest x-ray dataset containing three classes of annotations: lung, heart and clavicle. There are 247 chest radiographs in total, with unified resolution 2048$\times$2048. We split them into a training set of 197 images and a test set of 50 images, and resize them into 256$\times$256
\footnote{Clavicles are particularly small in chest x-ray images. To facilitate fine-grained segmentation and reduce consuming time, we crop their region of interest by statistics on the training set.}.

\noindent\textbf{Noise Patterns.}
To simulate manual noisy annotations, we randomly select a ratio $\alpha$ of samples from the training data to apply morphological~\cite{Zhu2019PickandLearnAQ, Zhang2020CharacterizingLE,Zhang2020RobustMI,Xue2020CascadedRL} or affine transformation with noise level controlled by $\beta$. For affine transformation, we use a combination of rotation and translation to imitate other real-world noise patterns. Unlike prior works, we use the relative size w.r.t the target object region when controlling the noise level $\beta$, as people usually annotate target object in a favorable field of view by zooming in or out images. 
We investigate our algorithm in several noisy settings with $\alpha$ being $\{0.3, 0.5, 0.7, 1.0\}$ and $\beta$ being $\{0.5, 0.7\}$. Some noisy examples are shown in the supplementary.

\subsection{Experiment Setup}   

\noindent\textbf{Comparisons} We compare our method with several state-of-the-art approaches, including Co-teaching~\cite{Han2018CoteachingRT}, Tri-network~\cite{Zhang2020RobustMI} and JoCoR~\cite{Wei2020CombatingNL}, which employ the robust learning at the pixel level. We do not include methods such as~\cite{ mirikharaji2019learning, Zhang2020CharacterizingLE} as they rely on a clean validation set. 
For fair comparison, we re-implement these methods with the same network backbone and training policy. 


\noindent\textbf{Implementation Details} We adopt nnU-Net~\cite{Isensee2020nnUNetAS} as the segmentation network. Following~\cite{Malach2017DecouplingT}, we take two networks sharing the same architecture yet with different initializations. 
Following~\cite{Han2018CoteachingRT}, the noise rate is assumed to be known, and we set {initial selection ratio} $\mathcal{R}$ as $(1 - \text{noise rate})$ and $\gamma$ as $1.1$. The balance factor $\lambda$ is 0.65.
We train our model by a SGD optimizer, with a constant learning rate 0.005. The batch sizes are 32 for ISIC dataset and 8 for JSRT dataset. We implement the code framework with PyTorch on TITAN Xp GPU.

\noindent\textbf{Evaluation Metric} During testing, we use the standard metric Dice coefficient (Dice) to evaluate the quality of predicted masks. We stop iterative learning when label refinement cannot bring any benefit, i.e., $G_l$ no longer shows the rising trend for training. To make fair comparisons, we train all methods for maximum 200 epochs and report average Dice over the last 10 epochs.

\begin{table}[t]
	\caption{Quantitative comparisons of noisy-labeled segmentation methods on ISIC dataset, where the metric is Dice[\%] over the last 10 epochs. $\alpha$ and $\beta$ control the noise ratio and noise level, respectively. } 
	\label{tab:isic}
	\centering
	\resizebox{\textwidth}{!}{
		\renewcommand{\arraystretch}{0.8}
		\begin{tabular}{P{2.5cm}|M{2.0cm}|M{2.3cm}|M{2.4cm}|M{2.2cm}|M{2.2cm}} 
			\toprule
			& Baseline  & Co-teaching\cite{Han2018CoteachingRT}   & Tri-network\cite{Zhang2020RobustMI}   & JoCoR~\cite{Wei2020CombatingNL}     & Ours \\ \midrule
			Original data           & 82.49  & 82.72  & 82.96  & 83.64  & \textbf{84.26}   \\ \midrule
			$\alpha=0.3, \beta=0.5$ & 80.75  & 81.44  & 81.50  & 82.65  & \textbf{84.00}  \\  \midrule
			$\alpha=0.3, \beta=0.7$ & 79.46  & 81.47  & 80.73  & 81.58  & \textbf{83.34}      \\  \midrule
			$\alpha=0.5, \beta=0.5$ & 78.95  & 81.22  & 80.94  & 82.41  & \textbf{83.90}      \\  \midrule
			$\alpha=0.5, \beta=0.7$ & 75.44  & 80.06  & 80.24  & 81.06  & \textbf{83.19}      \\  \midrule
			$\alpha=0.7, \beta=0.5$ & 76.61  & 79.61  & 79.55  & 80.55  & \textbf{83.83}      \\  \midrule
			$\alpha=0.7, \beta=0.7$ & 71.51  & 78.50  & 76.61  & 79.05  & \textbf{83.12}      \\  \midrule
			$\alpha=1.0, \beta=0.5$ & 71.13  & 76.69  & 75.61  & 78.43  & \textbf{82.23}      \\  \midrule
			$\alpha=1.0, \beta=0.7$ & 63.71  & 73.68  & 70.01  & 74.30  & \textbf{81.39}      \\  
			\bottomrule
		\end{tabular}
	}
\end{table}

\begin{figure}[t!]
	\centering 
	\includegraphics[width=1.00\textwidth]{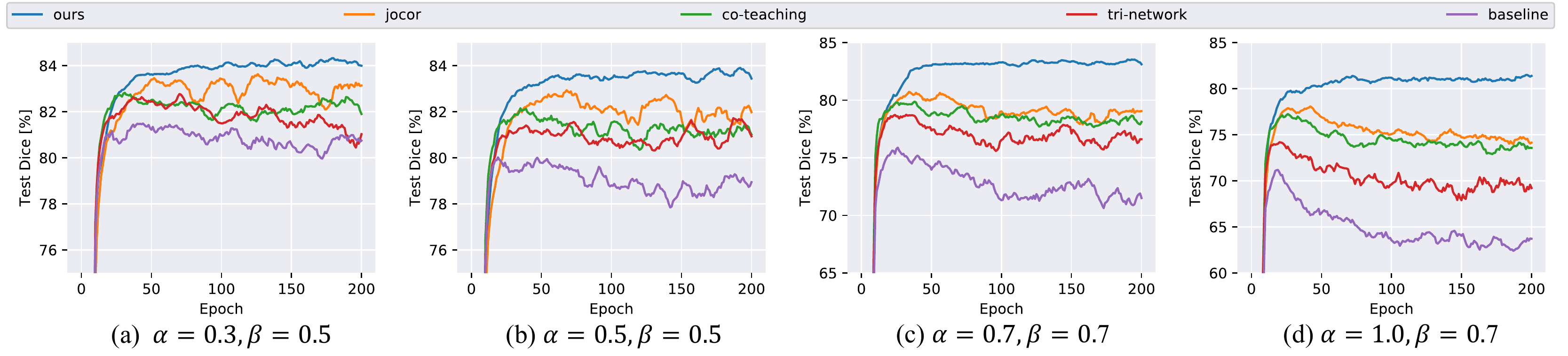}
	\caption{Curves of test dice vs. epoch on four different noise settings.} 
	\label{fig:test_dice}
\end{figure}

\subsection{Experiments on ISIC Dataset}



Table.~\ref{tab:isic} reports a summary of quantitative results of ISIC dataset.
At the mild noise setting $(\alpha=0.3, \beta=0.5)$, we achieve 84.00\% Dice and outperform recent methods more than 1.35\% Dice. As the noise increases, the performance of baseline decreases sharply, indicating the significant impact of label noise. Other methods mitigate this impact to some extent, but their performance still drop notably.
By contrast, our method consistently outperforms them and maintains high performance, validating its robustness to different noise settings. Remarkably, in the extreme noise setting ($\alpha=1.0, \beta=0.7$), our method achieves 81.39\% Dice and outperforms JoCoR (7.09\% Dice), Co-teaching (7.71\% Dice) and Tri-Network (11.38\% Dice)~\footnote{We also observe that our method outperforms the baseline with 84.26\% Dice on the original dataset, likely due to the noise in manual annotations~\cite{Gutman2018SkinLA}.}.

In Figure.~\ref{fig:test_dice}, we show curves of test dice vs. epochs. Most methods first reach a high performance then gradually decrease, indicating that their training is affected by noisy labels. In contrast, our method demonstrates a consistent high performance, which verifies the robustness of our training method. We also show some qualitative comparisons for visualization in the supplementary.



We also observe that our method is robust against the inaccuracy in superpixelization. For the ISIC dataset,  we use 100 superpixels per image with relative high undersegmentation error (1.0), and our superpixel selection can potentially discard inaccurate superpixels in the noise-aware learning. 

\subsection{Ablation Study}

We first verify the effect of superpixel representation based on a set of experiments on ISIC dataset under the noise setting $(\alpha=0.7, \beta=0.7)$, whose results are shown in Table.~\ref{tab:ablate}.
Row \#1 is our method which achieves 83.12\% Dice. Changing superpixel to pixel representation brings a performance drop of 1.97\% Dice in row \#2. This demonstrates the advantage of superpixel representation in learning with noisy labels.

In addition, to analyze the effect of selection module and label refinement module in iterative learning, we take a drop-one-out manner at the same setting. Ablating selection module from the model leads to a decrease of 3.80\% Dice in row \#3, meanwhile, removing label refinement module makes the performance drop 2.56\% Dice in row \#4. It is evident that both modules are essential for our robust iterative learning strategy. We also validate the effectiveness of adaptive stopping criterion and report the quality of refined labels in the supplementary.

\begin{table}[t!]
    \caption{Ablation study on our model components.}
    \label{tab:ablate}\small
	\centering
	\resizebox{0.825\textwidth}{!}{
        \setlength{\tabcolsep}{8pt}
        \renewcommand{\arraystretch}{0.9}
        \begin{tabular}[t]{c|c|c|c|c}
            \toprule
            Method  & Superpixel    & Selection    & Label Refinement    & Dice[\%]  \\ \midrule
            Ours    & $\checkmark$  & $\checkmark$  & $\checkmark$  &  \textbf{83.12} \\   
            		& $\times$  & $\checkmark$  & $\checkmark$      &  81.15 \\                 
                    & $\checkmark$  & $\times$  & $\checkmark$      &  79.32 \\
                    & $\checkmark$  & $\checkmark$  & $\times$      &  80.56 \\
            \bottomrule
        \end{tabular}
    }
\end{table}


\begin{figure*}[t!]
	\centering 
	\includegraphics[width=0.8\textwidth]{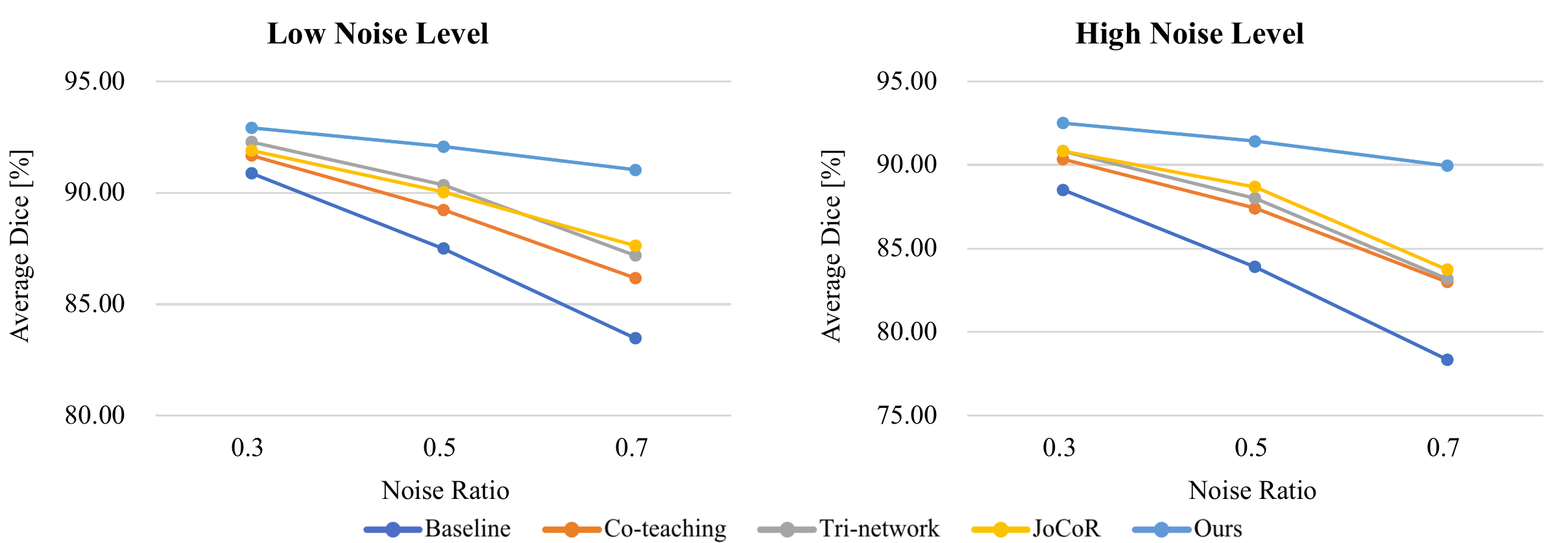}
	\caption{Average results on JSRT dataset with different noise settings: low noise level $\beta=0.5$ and  high noise level $\beta=0.7$, respectively. } 
	\label{fig:jsrt_curve}
\end{figure*}

\subsection{Experiments on JSRT Dataset}
To explore the generalization capability of our method, we also conduct experiments on JSRT dataset. Figure.~\ref{fig:jsrt_curve} presents the average results of three classes, and the table in supplementary reports detailed values for each class. Our method outperforms other methods consistently on all three classes. 

\section{Conclusion}
{
In this paper, we propose a robust learning strategy for medical image segmentation.
Unlike previous methods, we exploit structural prior and pixel correlation for segmentation model learning, which significantly mitigate the impact of label noise.
We develop an iterative learning scheme based on superpixel representation. In each iteration, we jointly train two deep networks using selected subsets of superpixels, and also relabel a subset of unreliable superpixels.
Evaluation on two benchmarks with simulated noises demonstrates that our learning strategy achieves the state-of-the-art performance and robustness in extensive noise settings. We note that learning with realistic label errors is an important future research topic and building a benchmark with such label noises is a crucial step.
}

\bibliographystyle{splncs04}
\bibliography{paper1575}

%
%




\end{document}


%
\title{Paper 1575 Supplementary}

%
\author{Shuailin Li $^*$ \inst{1} \and 
	Zhitong Gao $^*$ \inst{1} \and
	Xuming He\inst{1,2}}

%
\institute{
	ShanghaiTech University, Shanghai, China \\ \and
	Shanghai Engineering Research Center of Intelligent Vision and Imaging \\
	\email{\{lishl, gaozht, hexm\}@shanghaitech.edu.cn}
}

%
\maketitle              
%

\section{Noise patterns} 
\begin{figure*}[hb]
	\centering 
	\includegraphics[width=1.0\textwidth]{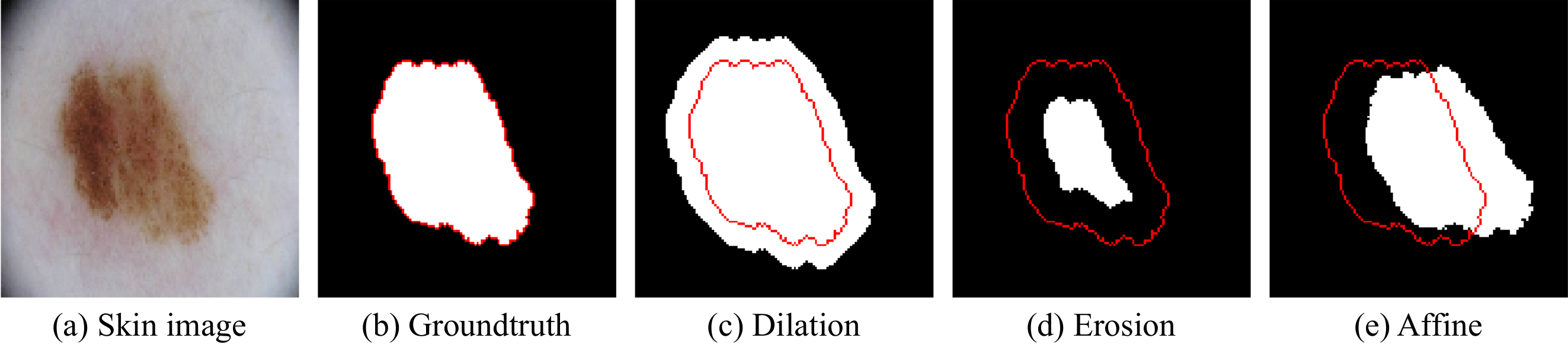}
	\caption{Visualization of noisy examples. Red curves indicate the groundtruth.}
	\label{fig:noise_pattern}
\end{figure*}

Figure.~\ref{fig:noise_pattern} shows three kinds of noise patterns: dilation, erosion and affine transformation.

\section{Visualization results} 
\begin{figure*}[t]
	
	\centering
	\begin{minipage}{0.9\linewidth}
		\captionsetup{font=footnotesize}
		\centering
		\subcaptionbox{ISIC dataset}
		{\includegraphics[width=1.00\textwidth]{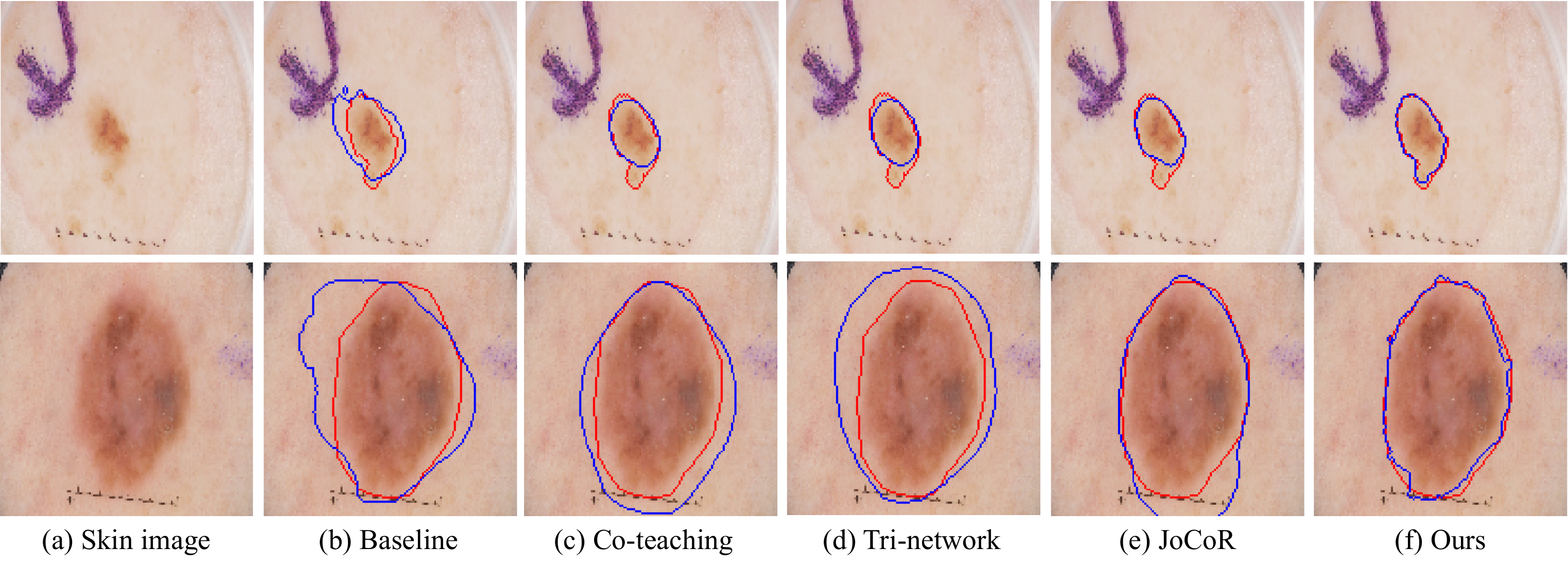}}
		
		\subcaptionbox{JSRT dataset}
		{\includegraphics[width=1.01\textwidth]{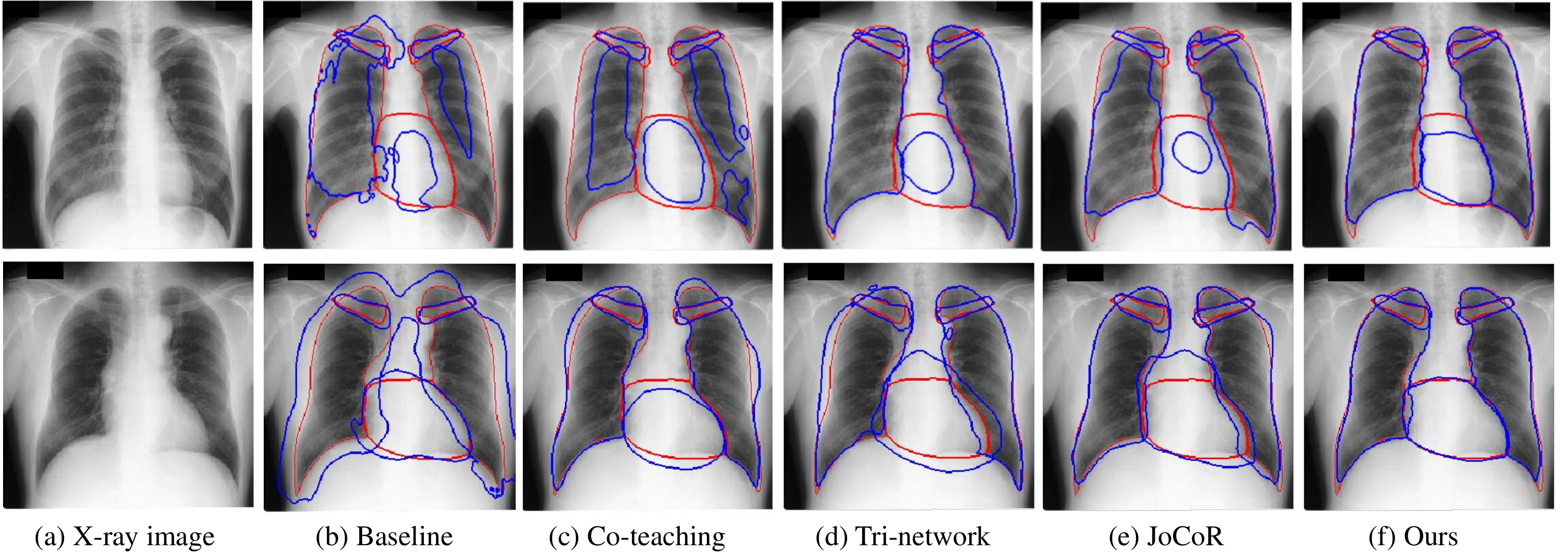}}
		
	\end{minipage}\quad
	\caption{Qualitative comparisons of different methods. Red and blue curves show the groundtruth and predicted masks, respectively.}
	\label{fig:vis}
	
\end{figure*}

We provide several qualitative results in Figure.~\ref{fig:vis} for visual comparison, in which models are trained under the noise setting $(\alpha=0.7, \beta=0.7)$.

\section{Ablation study on adaptive stopping} To validate the effect of our adaptive stopping criterion in iterative learning, we compare it with several fixed stopping epochs on ISIC dataset under the noise setting $(\alpha=0.7, \beta=0.7)$ in Table~\ref{tab:ablate_stop}. Results indicate its strong ability to prevent overfitting.

\begin{table}[t]
	\centering
	\caption{Ablation study on adaptive stopping criterion in iterative learning. Our criterion chooses epoch 26 here.}
	\label{tab:ablate_stop}
	\setlength{\tabcolsep}{6pt}
	\resizebox{1.0\textwidth}{!}{
		\begin{tabular}[t]{c|c|c|c|c|c|c}
			\toprule
			Stopping at & our criterion     & epoch 10 & epoch 20 & epoch 30 & epoch 50 & epoch 100   \\ \midrule  
			Dice [\%]        & \textbf{83.12}   & 81.58 & 82.38 & 80.78    & 80.75    & 80.78       \\          
			\bottomrule 
		\end{tabular}
	}
\end{table}

\section{The efficacy of the label refinement} We report the dice (\%) of final corrected labels on the training set in Table.~\ref{tab:label_refine}, which shows the improvement of our label refinement.

\begin{table}[t]
	\centering
	\caption{The efficacy of the label refinement. Each cell of the table displays the dice of corrected labels and dice of original noisy labels in the parenthesis.}
	\label{tab:label_refine}
	\setlength{\tabcolsep}{6pt}
	\resizebox{1.0\textwidth}{!}{
		\begin{tabular}[t]{c|c|c|c|c}
			\toprule
			Noise settings      &ISIC-lesion  &JSRT-Lung    &JSRT-heart   &JSRT-clavicle             \\ \midrule
			$\alpha=0.3,\beta=0.5$ & 92.99 (92.10) & 96.10 (91.73) & 94.21 (92.15) & 92.30 (92.02)    \\
			$\alpha=0.5,\beta=0.5$ & 90.25 (86.87) & 95.10 (86.23) & 93.38 (86.83) & 89.66 (86.58)    \\
			$\alpha=0.7,\beta=0.7$ & 84.07 (73.17) & 92.62 (73.05) & 89.30 (73.20) & 82.21 (73.29)    \\
			\bottomrule 
		\end{tabular}
	}
\end{table}

\section{Experiment on JSRT dataset}
We report quantitative results of JSRT dataset in Table.~\ref{tab:jsrt}. On all noisy settings of three classes, our method consistently outperforms other methods and show great robustness. 
Particularly, on the extreme noise setting $(\alpha=0.7, \beta=0.7)$, we achieve an Average Dice 89.95\%, outperforming JoCoR (83.73\%), Tri-network (83.18\%) and Co-teaching (82.99\%).

\begin{table}[hb]
	\caption{Quantitative comparisons of three classes on JSRT dataset.}
	\label{tab:jsrt}
	\centering
	\resizebox{1.0\textwidth}{!}
	{
		\renewcommand{\arraystretch}{0.9}
		\begin{tabular}{P{2.5cm}|M{2.0cm}|M{2.0cm}|M{2.0cm}|M{2.0cm}|M{2.0cm}} 
			\toprule
			Noise settings & Method    & Lung & Heart & Clavicle & Average     \\ \midrule
			Original data & Baseline & 97.64 & 94.04 & 90.89 & 94.19     \\ \midrule
			\multirow{5}{*}{$\alpha=0.3, \beta=0.5$}
			& Baseline  & 95.20  & 90.31  &87.12   & 90.88   \\ \cmidrule{2-6}
			& Co-teaching  & 95.57  & 92.27  &87.18   & 91.67   \\ \cmidrule{2-6}
			& Tri-network  & 96.20  & 92.59  &88.05   & 92.28   \\ \cmidrule{2-6}
			& JoCoR  & 95.67 & 92.68  &87.35   & 91.90  \\ \cmidrule{2-6}
			& Ours  & \textbf{96.87}  & \textbf{93.11}  &\textbf{88.77 }  &  \textbf{92.92}  \\ \midrule
			\multirow{5}{*}{$\alpha=0.3, \beta=0.7$}
			& Baseline  & 94.01  & 88.22  &83.29   & 88.51   \\ \cmidrule{2-6}
			& Co-teaching  & 94.75  &90.12   &86.15   & 90.34   \\ \cmidrule{2-6}
			& Tri-network  & 94.80  &90.90   &86.74   & 90.81   \\ \cmidrule{2-6}
			& JoCoR  & 94.94  &90.92   &86.61   & 90.82   \\ \cmidrule{2-6}
			& Ours  & \textbf{96.53}  &\textbf{92.52}   &\textbf{88.45}   &  \textbf{92.50}  \\ \midrule
			\multirow{5}{*}{$\alpha=0.5, \beta=0.5$}
			& Baseline  &89.72   &88.02   &84.75   & 87.50   \\ \cmidrule{2-6}
			& Co-teaching  &91.93   &88.78   &86.99   & 89.23   \\ \cmidrule{2-6}
			& Tri-network  & 93.09  &90.64   &87.32   & 90.35   \\ \cmidrule{2-6}
			& JoCoR  & 92.74  &90.57    &86.80   & 90.04  \\ \cmidrule{2-6}
			& Ours  &\textbf{96.24 }  &\textbf{92.27}   &\textbf{87.72}   &  \textbf{92.08}  \\ \midrule
			\multirow{5}{*}{$\alpha=0.5, \beta=0.7$}
			& Baseline  &87.46   &81.26   &82.96   & 83.89   \\ \cmidrule{2-6}
			& Co-teaching  &89.70   &87.28   &85.24   & 87.41   \\ \cmidrule{2-6}
			& Tri-network  &89.87   &88.16   &85.99   & 88.01   \\ \cmidrule{2-6}
			& JoCoR  &91.27   &88.86   &85.94   & 88.69   \\ \cmidrule{2-6}
			& Ours  &\textbf{95.87 }  &\textbf{91.76}   &\textbf{86.65}   &  \textbf{91.43}  \\ \midrule
			\multirow{5}{*}{$\alpha=0.7, \beta=0.5$}
			& Baseline  &83.71   &83.62   &83.06   & 83.46   \\ \cmidrule{2-6}
			& Co-teaching  &86.45   & 86.21  &85.85  & 86.17   \\ \cmidrule{2-6}
			& Tri-network  &87.75   & 87.46   &86.34   & 87.18   \\ \cmidrule{2-6}
			& JoCoR  &89.22   &88.33   &85.32   & 87.62   \\ \cmidrule{2-6}
			& Ours  &\textbf{94.93 }  &\textbf{91.49}   &\textbf{86.68}   &  \textbf{91.03}   \\ \midrule
			\multirow{5}{*}{$\alpha=0.7, \beta=0.7$}
			& Baseline  &79.82   &78.12   &77.09   & 78.34   \\ \cmidrule{2-6}
			& Co-teaching  &84.71   &82.01   &82.24   & 82.99   \\ \cmidrule{2-6}
			& Tri-network  &84.35   &82.48   &82.70   & 83.18   \\ \cmidrule{2-6}
			& JoCoR  &85.94   &83.91   &81.33   & 83.73   \\ \cmidrule{2-6}
			& Ours  &\textbf{94.59}   &\textbf{90.20}   &\textbf{85.07}   &  \textbf{89.95}  \\ 
			\bottomrule
		\end{tabular}
	}
\end{table}

\bibliographystyle{splncs04}

%
%



